# Quantum Machine Learning in the Cognitive Domain: Alzheimer's Disease Study

Emine AKPINAR
Department of Physics
Yildiz Technical University
Istanbul, Turkey

*Abstract*— Alzheimer's disease (AD) is the most prevalent neurodegenerative disorder, primarily affecting the elderly population and leading to significant cognitive decline. This decline manifests in various mental faculties such as attention, memory, and higher-order cognitive functions, severely impacting an individual's ability to comprehend information, acquire new knowledge, and communicate effectively. One of the tasks influenced by cognitive impairments is handwriting. By analyzing specific features of handwriting, including pressure, velocity, and spatial organization, researchers can detect subtle changes that may indicate early-stage cognitive impairments, particularly AD. Recent developments in classical artificial intelligence (AI) methods have shown promise in detecting AD through handwriting analysis. However, as the dataset size increases, these AI approaches demand greater computational resources, and diagnoses are often affected by limited classical vector spaces and feature correlations. Recent studies have shown that quantum computing technologies, developed by harnessing the unique properties of quantum particles such as superposition and entanglement, can not only address the aforementioned problems but also accelerate complex data analysis and enable more efficient processing of large datasets. In this study, we propose a variational quantum classifier with fewer circuit elements to facilitate early AD diagnosis based on handwriting data. Our model has demonstrated comparable classification performance to classical methods and underscores the potential of quantum computing models in addressing cognitive problems, paving the way for future research in this domain.

## I. INTRODUCTION

Cognitive impairments encompass a diverse range of neurological conditions characterized by varied clinical and pathological manifestations, targeting distinct subsets of neurons within specific functional anatomical networks [1]. Cognitive impairment can range from loss of function in multiple mental abilities: concentration, memory, attention, language, problem-solving, decision-making, and other higher-order thinking abilities. These impairments can affect a person's ability to understand information, learn new things, communicate effectively, and perform routine tasks. Although their symptoms can remain unchanged or even vanish, for the majority of patients, the condition evolves into dementia diseases. Therefore, early diagnosis can assist both the patient and caregivers in better managing the coping process with the disease and contribute to the patient's ability to maintain their daily life activities more effectively [2]. Furthermore, being aware of the severity and progression risks enables patients to take precautions before irreversible brain damage occurs. One of the most common conditions associated with cognitive impairment is Alzheimer's disease (AD) [3, 4]. AD is a neurodegenerative, progressive brain disorder that typically manifests in old age and is one of the leading causes of dementia in today's world. According to the World Alzheimer's Report (2018) [5], approximately 50 million people were affected by this disease in 2018, and it is expected to triple by the year 2050. AD usually begins with mild symptoms and may eventually lead to a severe stage where the patient has difficulty in physical abilities and loses awareness [6].

Criteria for clinical diagnosis of AD were proposed in 1984 [7] by the National Institute of Neurological and Communicative Disorders and Stroke (NINCDS) and by the Alzheimer's Disease and Related Disorders Association (ADRDA). According to these criteria, the diagnosis of AD needs histopathologic confirmation in autopsy or biopsy. To date, in addition to these, clinical diagnosis of such diseases was supported by tools such as imaging, cognitive tests, blood tests…etc. Nonetheless, an early and accurate diagnosis would greatly improve the effectiveness of available treatments, but it is still a challenging task.

In recent studies, it has been demonstrated that individuals with AD diseases demonstrate changes in spatial organization and impaired motor control [8]. Therefore, some diagnostic signs of AD should be detectable through motor tasks. It has been demonstrated that handwriting analysis can provide valuable insights into the cognitive decline impairment with AD [9]. Handwriting is the result of a complex network of cognitive, kinesthetic, and perceptual-motor skills. AD can disrupt the coordination and control of these skills, leading to significant changes in handwriting. By analyzing various aspects of handwriting, such as pressure, velocity, and spatial organization, researchers can identify subtle changes that may indicate cognitive impairments in its early stages [10].

However, in this field, many published studies have been conducted in the areas of medicine and psychology, where typically the relationships between disease and handwriting variables have been determined using classical statistical tools (including ANOVA and MANOVA analysis) [11]. Studies in literature utilize machine learning and deep learning-based classification methods to detect people affected by AD using handwriting information [12,13]. However, advanced classical artificial intelligence (AI) methods require higher computational power as the size of the handwriting data increases. Additionally, the diagnoses are influenced by factors such as the noise level of the data, low resolution, limited relevant classical vector space, and correlations between features [14].

Recent studies have shown that the use of quantum computing technologies in the healthcare field can not only solve these problems but also accelerate complex data analysis and process large datasets more efficiently [15]. Despite the limitations of NISQ (Noisy Intermediate-Scale Quantum) computers, numerous studies have demonstrated the potential for quantum computers to outperform classical computers in certain AI applications.

To the best of our knowledge, this is the first study in the cognitive field to utilize quantum computing methods. In this study, we introduce a novel variational quantum classifier that achieves higher classification accuracy with fewer circuit elements, making it fully compatible with NISQ (Noisy Intermediate-Scale Quantum) computers. Our goal with this model is to detect AD using handwriting data.

## II. MATERIALS AND METHODS

There are three main stages in the proposed study: data collection and pre-processing, classification using the proposed variational quantum classifier, and post-processing.

### A. Dataset and Pre-Processing

The effectiveness of the proposed variational quantum classifier is evaluated on real AD patient dataset which are taken from UCI [16, 17]. In this dataset, tests were conducted on 174 patients with AD according to specific protocols. In these tests, the cognitive abilities of the examined subject were assessed by using questionnaires including questions and problems in many areas, which range from orientation to time and place, to registration recall. According to the protocol, nine tasks were introduced to the patients, and the handwriting dynamics of the patients were recorded. Following the determination of the categorical values in the data set, these values were converted to numerical values using one-hot encoding method.

After obtaining the dataset, the following preprocessing steps were performed:

*a) Principal Component Analysis (PCA):* Due to the limited number of qubits in real quantum computers, dimensionality reduction is required before mapping features to the quantum space. In PCA, the goal is to find the principal components, which are new orthogonal axes that represent the directions of maximum variance in the data [18]. Keeping the data variance ensures that the quintessential patterns in the data are preserved even though the dimensionality is (dramatically) reduced. PCA uses methods like singular value decomposition to find a linear transformation to a new set of coordinate axes (a new basis of the vector space), such that the variance along each of the new axes is maximized. In this study, obtained principal components of the dataset must match the number of qubits our circuit has.

*b) Normalization:* Data normalization, especially in quantum machine learning, enables data to be consistently transformed into quantum states, thereby improving the efficiency of quantum algorithms. In this study, the Min-Max normalization method is aimed to be used. Min-Max normalization is a data normalization technique that scales data features to a specific range while preserving their original distributions. This method brings data features to a certain minimum and maximum range. Min-Max normalization is usually applied using the following formula:

$$x_{normalize} = \frac{(x - \min(x))}{\max(x) - \min(x)} \quad (1)$$

In here, $x_{normalize}$, represents the normalized value, $x$, represents the original value, $\min(x)$, is the minimum value and $\max(x)$, is the maximum value of the data feature.

### B. Variational Quantum Classifier

Variational quantum classifier (VQC) is a supervised quantum machine learning algorithm that enables experimental results to be obtained on NISQ quantum computers without the need for error correction techniques. The algorithm uses a hybrid approach where the parameters are optimized and updated on a classical computer, allowing the optimization process to be carried out without increasing the required coherence times [19].

VQC involves three main steps:

- initial state preparation with feature mapping methods,
- building a parametrized quantum circuit or variational ansatz for the classification task,
- the measurement stage to assess class probabilities.

The optimization process is performed on a classical computer to update the parameters of the variational ansatz [20]. Thus, this process is not a part of quantum variational circuit. The general quantum circuit structure of the VQC model used in our study is shown in Figure 1. The circuit includes the feature map, parameterized quantum circuit, and measurement stages, presented in sequence.

*a) Feature Map:* Quantum computers only interact with data expressed as quantum states. Therefore, classical data must be transformed into quantum states to be processed by quantum algorithms. Feature mapping methods used to embed classical data from classical vector space to potentially vastly higher-dimensional feature space, quantum Hilbert space (in lower dimension, quantum state space) with non-linearly. In addition, mapping two data inputs into a feature space and taking their inner product leads to a kernel function that measures the distance between the data points.

Feature maps are utilized to enhance the learning algorithm's performance and achieve more accurate predictions. Mathematically, the quantum feature mapping function maps classical data points, $\vec{x}$ input, non-linearly to a quantum state, realizes the map:

$$\phi: \vec{x} \mapsto |\phi(\vec{x})\rangle\langle\phi(\vec{x})| \quad (2)$$

$\phi(\vec{x})$ represents the quantum feature map. In this study, we used ZZFeatureMap which is second order expansion of Pauli feature map (($U_{\varphi(\vec{x})}$) [21, 22]:

$$U_{\varphi(\vec{x})} = exp\left(i \sum_{S \subseteq [n]} \varphi_S(\vec{x}) \prod_{i \in S} P_i\right). \quad (3)$$

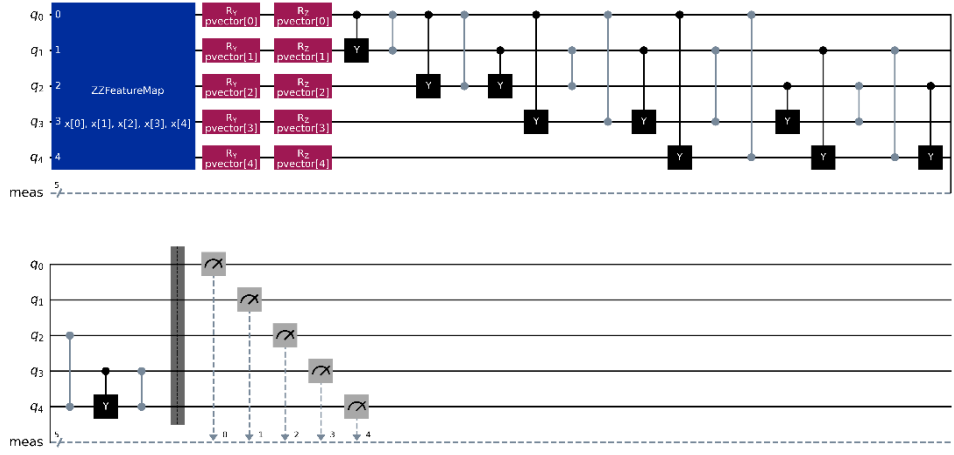

**Figure 1:** This figure shows the overall quantum circuit structure of the VQC model used in our study. The circuit was drawn after applying only one repetition of the model.

where $P_i$ denote the Pauli matrices, $P_i \in \{I, X, Y, Z\}$ and $S$ define the connectivity between different qubits. If we customize the Pauli gates as $P_0 = X$, $P_1 = Y$, $P_2 = ZZ$, the ZZFeatureMap is obtained as follows:

$$U_{\varphi(\vec{x_i})} = \begin{pmatrix} exp\left(i \sum_{jk} \varphi_{\{j,k\}}(\vec{x}) Z_j \otimes Z_k\right) \\ exp\left(i \sum_{j} \varphi_{\{j\}}(\vec{x}) Y_j\right) \\ exp\left(i \sum_{j} \varphi_{\{j\}}(\vec{x}) X_j\right) H^{\otimes n} \end{pmatrix}^d \quad (4)$$

The ZZFeatureMap has a structure that is difficult to simulate classically due to its entangling blocks. This characteristic is crucial for gaining a computational advantage over classical approaches.

*b) Parametrized Quantum Circuit or Variational Ansatz:* The variational ansatz, also known as the parameterized quantum circuit (PQC), is a unitary transformation composed of a set of one- or two-qubit quantum gates with tunable parameters (θ). It performs operations on quantum states [23]. Essentially, quantum devices with these tunable parameters allow for adjustments in the computational specifications. The key insight is that the circuit parameters can be optimized until the PQC produces the desired output. The parameters of the quantum gates, including single-qubit gates, are adjusted using classical iterative optimization processes [24]. This component forms the basis for the training and testing processes in our study.

The quantum circuit model $U(\theta)|\phi(x)>$ transforms a quantum state $\psi(x, \theta) >$ by applying a set of operations. Here, $U(\theta)$, consists of a set of one- and two-qubit unitary gates. In other words, transformations that map quantum states to other quantum states, known as quantum gates, must be unitary. Instead of $U(\theta)$, the expression Rot $(\theta_1, \theta_2, \theta_3)$ can also be used, where $\theta_1, \theta_2, \theta_3$ are parameters that can be optimized during the optimization process [25].

$$|\psi(x,\theta)> = \text{Rot}(\theta_1,\theta_2,\theta_3) |\phi(x)> =$$
$$\begin{pmatrix} e^{i\left(-\frac{\theta_1}{2}-\frac{\theta_3}{2}\right)}\cos\left(\frac{\theta_2}{2}\right)\cos\left(\frac{x}{2}\right) + ie^{i\left(-\frac{\theta_1}{2}+\frac{\theta_3}{2}\right)}\sin\left(\frac{\theta_2}{2}\right)\sin\left(\frac{x}{2}\right) \\ e^{i\left(\frac{\theta_1}{2}-\frac{\theta_3}{2}\right)}\sin\left(\frac{\theta_2}{2}\right)\cos\left(\frac{x}{2}\right) - ie^{i\left(\frac{\theta_1}{2}+\frac{\theta_3}{2}\right)}\cos\left(\frac{\theta_2}{2}\right)\sin\left(\frac{x}{2}\right) \end{pmatrix} \quad (5)$$

The PQC employs $R_Y$ and $R_Z$ gates for parameterization, which are interlinked through a set of parameters. The circuit also incorporates entanglement using $C_Y$ and $C_Z$ gates to create the necessary quantum correlations.

$$R_Y(\theta) = \begin{bmatrix} \cos\left(\frac{\theta}{2}\right) & -\sin\left(\frac{\theta}{2}\right) \\ \sin\left(\frac{\theta}{2}\right) & \cos\left(\frac{\theta}{2}\right) \end{bmatrix},$$

$$R_Z(\theta) = \begin{bmatrix} e^{-i\frac{\theta}{2}} & 0 \\ 0 & e^{i\frac{\theta}{2}} \end{bmatrix}, \quad (6)$$

$$C_Y = \begin{bmatrix} 1 & 0 & 0 & 0 \\ 0 & 1 & 0 & 0 \\ 0 & 0 & 0 & -i \\ 0 & 0 & i & 0 \end{bmatrix}, \quad (7)$$

$$C_Z = \begin{bmatrix} 1 & 0 & 0 & 0 \\ 0 & 1 & 0 & 0 \\ 0 & 0 & 1 & 0 \\ 0 & 0 & 0 & -1 \end{bmatrix}.$$

*c) Measurement:* After performing training and testing processes with the variational ansatz circuit, a measurement function is applied to learn the predicted class labels and obtain class probabilities. To stabilize the stochastic effects of the measurements, the quantum circuit will be measured a number of times to estimate the probability of its outcomes. In fact, we run our proposed circuit 1024 shots. In our study, for AD binary classification task, the results are measured in Paulie Z basis. The binary measurement is obtained from the parity function $f = Z_1 Z_2$.

The results, which are bit strings $z \in \{0,1\}^2$ assigned to a label based on a predetermined boolean function $f: \{0, 1\} \to \{-1, 1\}$.

By computing the parity of the bit string,

- if the measurement consists of an even number of '1's, it will be mapped to the AD class,
- if the measurement outcome consists of an odd number of '1's, it is mapped to the non-AD class.

This means that the class label is determined based on whether the number of '1's in the measurement outcome is even or odd.

*d) Optimization*: To improve prediction results, the parameters of the single-qubit quantum gates in the PQC need to be updated iteratively to find their optimal values. In the variational quantum circuit, the optimization process is illustrated in Figure 2.

The study employed the SPSA (Simultaneous Perturbation Stochastic Approximation) optimization technique. SPSA utilizes simultaneous perturbations to estimate the gradient of the loss function and then updates the parameters in the direction that improves the loss function [26]. SPSA features various methods designed to find the global minimum

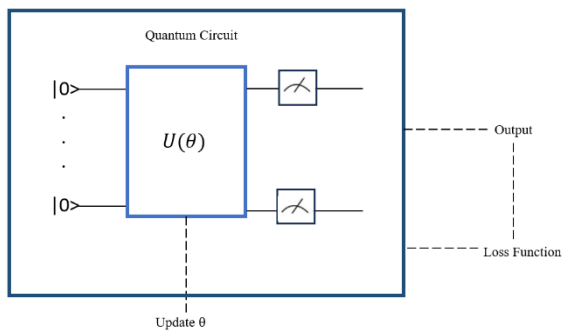

**Figure 2.** Optimization process in variational quantum circuits.

*C. Evaluation*

To evaluate the performance of the proposed variational quantum classifier, the area under receiver operating characteristic curve (AUROC) analysis was utilized along with accuracy, sensitivity, specificity, and F1-score metrics obtained from the confusion matrix. The AUROC curve helps in assessing the trade-off between true positive rate (sensitivity) and false positive rate. The confusion matrix provides more detailed information on how well the prediction model is performing, including true positives (TP), true negatives (TN), false positives (FP), and false negatives (FN).

These elements help to determine which classes are correctly classified and which ones are misclassified.

In addition to these metrics, kernel matrices for both the training and testing datasets obtained using ZZFeatureMap. Feature maps provide insights into which features contribute to misclassifications, allowing for a more detailed analysis of the classifier's performance. Overall, these evaluation metrics and feature map will help understand the classifier's strengths and weaknesses and provide valuable insights for further improvements.

$$Accuracy = \frac{T_P + T_N}{T_P + T_N + F_P + F_N}$$

$$Sensitivity = \frac{T_P}{(T_P + F_N)}$$

$$Specificity = \frac{T_N}{(T_N + F_P)}$$

$$F1 - score = \frac{2 * T_P}{(2 * T_P + F_P + F_N)}$$

(8)

### III. RESULTS

The performance of our proposed variational quantum classifier on the Alzheimer's patients' handwriting dataset is depicted in Table 1. The values of TP, TN, FP, and FN for the variational quantum classifier are presented in Figure 3. Each of these values is available in the confusion matrix for both AD and Non-AD cohorts.

As evident from Table 1, for the AD cohort, the performance metrics include an accuracy of 0.75, specificity of 0.69, sensitivity of 0.88, F1-score of 0.77, and an AUROC value of 0.68. The corresponding values for the non-AD cohort are 0.75 for accuracy, 0.75 for specificity, 0.50 for sensitivity, 0.59 for F1-score, and 0.68 for AUROC.

In comparison, the performance of various machine learning algorithms on the same dataset is summarized as follows: the K-Nearest Neighbors (KNN) model achieved an accuracy of 0.74, the Support Vector Machine (SVM) algorithm performed the best with an accuracy of 0.79, and the Decision Tree (DT) model attained an accuracy of 0.75.

|  | Accuracy | Specificity | Sensitivity | F1-Score | AUROC |
|---|---|---|---|---|---|
| AD cohort | 0.75 | 0.69 | 0.88 | 0.77 | 0.68 |
| Non-AD cohort | 0.75 | 0.75 | 0.50 | 0.59 | 0.68 |

**Table 1.** The table presents a comprehensive set of evaluation metrics, including accuracy, specificity, sensitivity, F1-score and AUROC to assess the proposed variational quantum classifier efficacy and performance.

Due to the availability of cloud-based backends in IBMQ, the implementation of the proposed variational quantum classifier was carried out using IBM's Qiskit version 0.36.1 framework. Data classification was performed using the related simulator Aer version 0.12.2. We used ibmq qasm simulator, which is a 32 qubit simulator.

In this study, the SPSA optimizer was utilized with a maximum of 500 iterations. In Figure 4, the loss function shows a consistent downward trend. Although the graph exhibits some spikes, it is understandable since we are utilizing stochastic gradient-descent, and there is a possibility of randomness present.

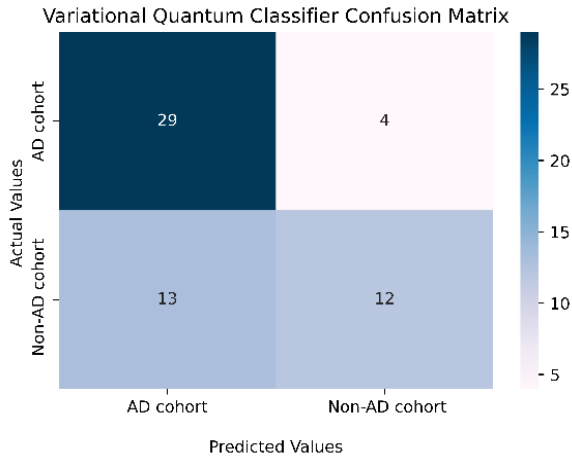

**Figure 3.** Confusion matrix of proposed variational quantum classifier.

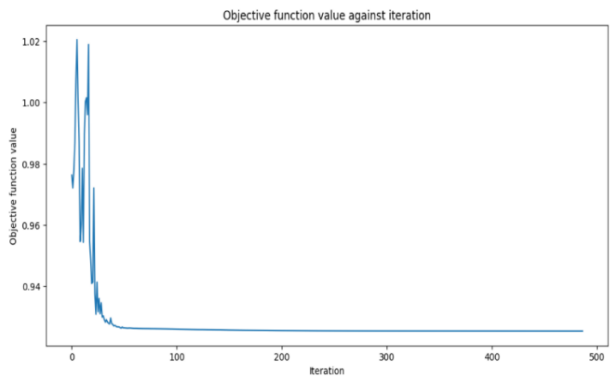

**Figure 4.** The objective function value of the proposed model against the selected number of iterations.

Figure 5 illustrates the construction of a kernel matrix that contains the inner product of all data points used for training and testing. This enables us to invoke the classical kernel function from the training and testing data in the feature map.

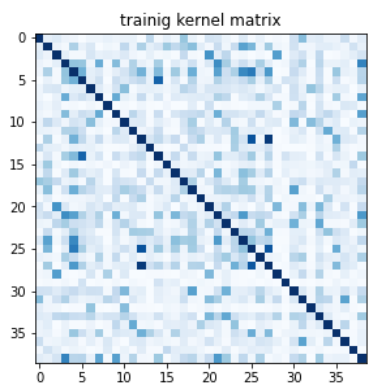

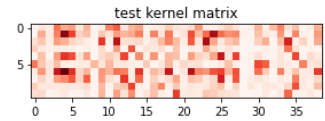

**Figure 5.** Kernel matrices for both training and testing phases were generated by the ZZFeatureMap.

## IV. CONCLUSION

Although current NISQ quantum computers have a limited number of qubits, it is anticipated that future quantum computers will have a substantial impact across various data processing domains, particularly in big data, where datasets are pushing the limits of classical computational resources. Medical image analysis, along with obtaining actionable insights from such data, has also gained significant importance in recent years. In this context, a wide range of quantum applications is being explored in the medical field.

In this study, we employed the proposed variational quantum classifier to aid in the early diagnosis of AD through handwriting analysis. Our results demonstrated that the proposed classifier achieved a classification accuracy of 0.75 in distinguishing the AD cohort from the non-AD cohort. Furthermore, when compared to the performance of several widely-used machine learning algorithms, we have shown that proposed model can offer comparable accuracy to classical methods in early diagnosing of AD based on cognitive changes.

In future studies, we aim to further refine the proposed variational quantum classifier and achieve higher diagnostic accuracy for AD, focusing on the integration of more complex cognitive indicators and enhanced quantum algorithm.